\documentclass{article}

\usepackage{arxiv}

\usepackage[utf8]{inputenc} 
\usepackage[T1]{fontenc}    
\usepackage{hyperref}       
\usepackage{url}            
\usepackage{booktabs}       
\usepackage{amsfonts}       
\usepackage{nicefrac}       
\usepackage{microtype}      
\usepackage{lipsum}
\usepackage{graphicx}
\usepackage{subcaption}
\usepackage{amsmath, amsthm, amssymb, amsfonts, amsbsy}
\usepackage{mathtools}
\usepackage{caption}
\usepackage{wrapfig}
\usepackage{xcolor}
\usepackage{float}

\newcommand*{\ora}{\overrightarrow}

\title{ Scaled-Time-Attention Robust Edge Network}

\author{ 
  Richard Lau \\
	Peraton Labs \\
	Red Bank, NJ 07701 \\ 
	\texttt{clau@peratonlabs.com} \\
	\And
  Lihan Yao \\
	Peraton Labs \\
	Red Bank, NJ 07701 \\ 
	\texttt{lihan.yao@peratonlabs.com} \\
	\And
	Todd Huster \\
	Peraton Labs \\
	Red Bank, NJ 07701 \\ 
	\texttt{thuster@peratonlabs.com} \\
	\And
	William Johnson \\
	Peraton Labs \\
	Red Bank, NJ 07701 \\ 
	\texttt{wjohnson@peratonlabs.com} \\
	\And
	Stephen Arleth \\
	Peraton Labs \\
	Red Bank, NJ 07701 \\ 
  \texttt{sarleth@peratonlabs.com} \\
  \And 
  Justin Wong \\
  \texttt{jjwong997@gmail.com} \\
	\And
	Devin Ridge \\
	Virginia Tech \\
  Blacksburg, VA 24061 \\
	\texttt{devinr@vt.edu} \\
	\And
	Michael Fletcher \\
	Virginia Tech \\
  Blacksburg, VA 24061 \\
	\texttt{mjf@vt.edu} \\
	\And
	William C. Headley \\
  Virginia Tech \\
  Blacksburg, VA 24061 \\
	\texttt{cheadley@vt.edu} 
  }
  
\begin{document}
\maketitle

\begin{abstract}
  This paper describes a systematic approach towards building a new family of neural networks based on a delay-loop version of a reservoir neural network. The resulting architecture, called Scaled-Time-Attention Robust Edge (STARE) network,  exploits hyper dimensional space and non-multiply-and-add computation to achieve a simpler architecture, which has shallow layers, is simple to train, and is better suited for Edge applications, such as Internet of Things (IoT), over traditional deep neural networks. STARE incorporates new AI concepts such as Attention and Context, and is best suited for temporal feature extraction and classification. We demonstrate that STARE is applicable to a variety of applications with improved performance and lower implementation complexity. In particular, we showed a novel way of applying a dual-loop configuration to detection and identification of drone vs bird in a counter Unmanned Air Systems (UAS) detection application by exploiting both spatial (video frame) and temporal (trajectory) information.  We also demonstrated that the STARE performance approaches that of a State-of-the-Art deep neural network in classifying RF modulations, and outperforms Long Short-term Memory (LSTM) in a special case of Mackey Glass time series prediction. To demonstrate hardware efficiency, we designed and developed an FPGA implementation of the STARE algorithm to demonstrate its low-power and high-throughput operations. In addition, we illustrate an efficient structure for integrating a massively parallel implementation of the STARE algorithm for ASIC implementation. 
\end{abstract}


\section{Introduction}
\label{sec:headings}

\graphicspath{{/home/lyao/paper_hyddenn/figs/}}

\begin{wrapfigure}{L}{0.55\textwidth}
  \centering
  \includegraphics[width=9cm,height=4cm]{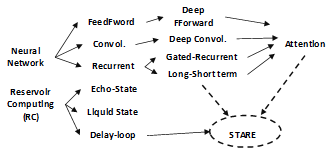} 
  \captionof{figure}{STARE relationship with other neural networks.}
  \label{fig:fig1}
\end{wrapfigure}

Deep Neural Networks (DNN) have been successful in the past few years. While its applications were initially found to be effective in image classification, object recognition, and Natural Language Processing, they have been recently extended to other areas including RF modulation classification, Dynamic Spectrum Access, system identification, and time series prediction. As the applications of DNN proliferate in many different domains, such as Internet of Things (IoT), 5G and 6G mobile networks, power grid networks, and cyber security, the issue of complexity, scalability, miniaturization, and hardware implementation become key research topics. These new requirements suggest new investigation in alternative neural network architectures where complexity and scalability becomes part of the optimization criteria. To address these new requirements, we propose Scaled-Time-Attention Robust Edge (STARE) network, which is part of a study on a DARPA1 Artificial Intelligence Exploratory (AIE) program on Hyper Dimensional Data Enabled Neural Network (HyDDENN) 
\cite{darpa1}.    

STARE belongs to the category of Reservoir Computing (RC) neural networks. RC originally evolved from Recurrent Neural Networks (RNN), but have a shallow architecture that consists of a reservoir with sparse random weights followed by a linear regression readout layer. RC works by projecting the input signal into a hyper-dimensional space, followed by a training layer via linear regression. In recent years, a new RC architecture emerged in the form of a delay-loop reservoir (DLR). The DLR replaces the traditional RC reservoir, which is a mesh network with a single delay loop. This new architecture was found to exhibit similar performance as the traditional RC but has the important benefit of being suitable for hardware implementation such as using photonic fiber for the delay loop 
\cite{darpa2, van2017advances3, lukovsevivcius2009reservoir4}.  

Although the DLR has advanced the neural network technology by a simplified reservoir structure, a single delay loop reservoir cannot control the temporal features. Conceptually, the DLR does not know how long should the reservoir “remember” the past, and thus does not have the concept of a context. Note that these temporal scaling and contextual concepts were also missing in the traditional RC or NN architectures. In recent years, these deficiencies have been addressed in LSTM (Long short-term memory) neural networks and attention networks. As illustrated in Figure 1, the STARE architecture incorporates the key concepts of LSTM and attention networks, and yet retains the desirable characteristics of the shallow architecture, hyper-dimensional spatial-temporal expansion, and simplified loop-based reservoir structure of the DLR. The resulting STARE network can be implemented in low-power, low-precision computing, FPGA platforms. Future development with ASIC implementation would be feasible for incorporation into next generation edge devices.
\let\thefootnote\relax\footnotetext{DARPA DISTRIBUTION STATEMENT A. Approved for public release. Distribution Unlimited.}

\section{STARE Architecture}
\label{sec:architecture}

The STARE architecture is based on a delay loop version of Reservoir Computing (RC), which performs nonlinear transformation of the incoming signal into hyper dimensional space to separate nearby or tangled signals, followed by a simple linear classifier. A major difference between RC and Recursive Neural network (RNN) is that the reservoir in RC has a static structure (fixed weights and interconnections) and does not require training, while that of the RNN requires training. Moreover, it was found that performance is not sensitive to adding more connections in the reservoir, and delay loops are also found to be more stable than RNN.

In STARE, the reservoir is implemented as one or more delay loops. Different delay-loop reservoir configurations depend on how the loops are combined together. Specifically, the following configurations are found to be versatile:

\begin{enumerate}
  \item[1] This configuration has one or multiple delay loops followed by a linear regressor. Refer to a single delay loop configuration as Single-loop Reservoir, or SLR. The N parallel delay loops configuration is called Multiple-Loop Reservoir, or MLR.
  \item[2] The second configuration is called  delay loops (DL-Tree), which has a tree structure where each node of the tree is a SLR or MLR.
\end{enumerate}

These different configurations give different results depending how it is used. For simple applications such as the RF emitter classification, the SLR with small hyper dimension (N) is sufficient. For applications where temporal features are important, such as in Mackley Glass signal prediction and counter-UAS detection, a MLR proves to be more effective. In demanding applications such as classifying a large number of bursty signals (classification of DeepSig RF modulation), the DL-tree configuration offers the best result. In the following, we describe in detail how a dual-loop MLR is configured, with which the configuration of SLR and general MLR follow.

\begin{figure}[h!]
  \centering
  \includegraphics[width=15cm,height=6cm]{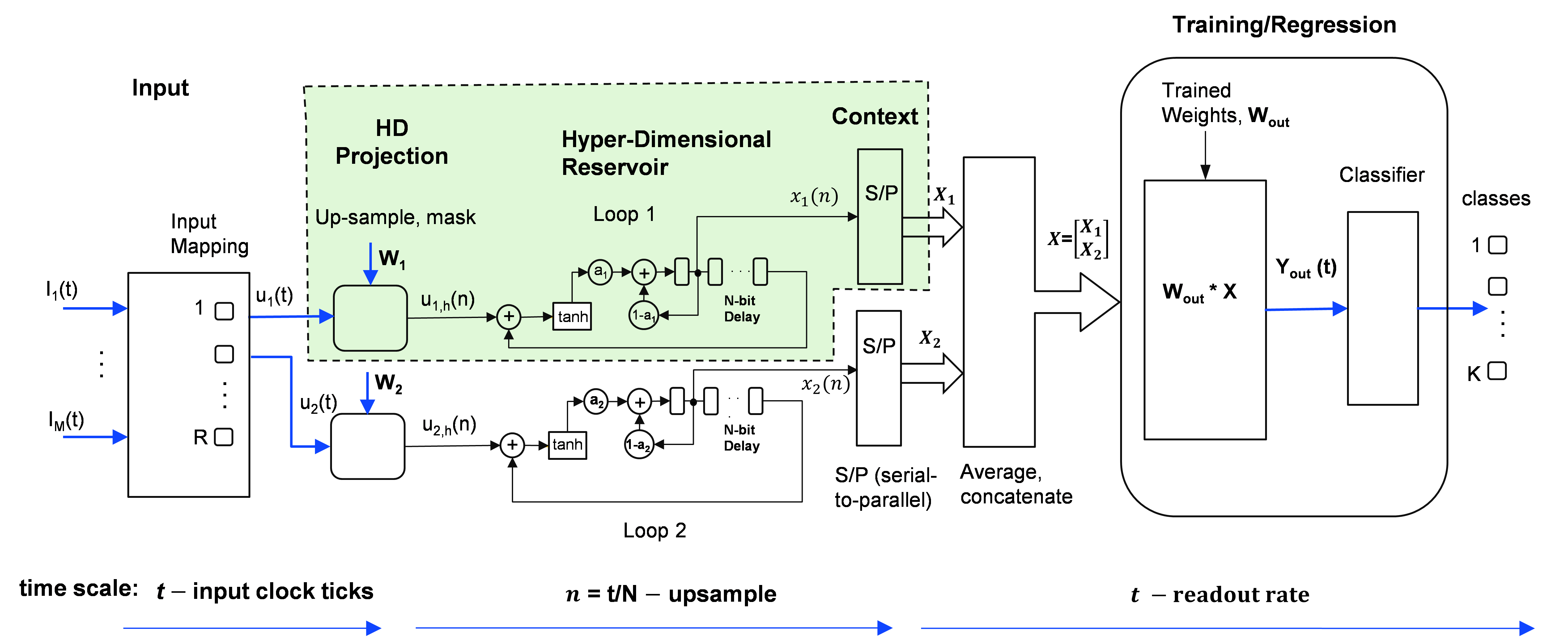}
  \captionof{figure}{STARE dual-loop MLR architecture.}
  \label{fig:fig2}
\end{figure}

\subsection{Dual-loop MLR}

Conceptually, the STARE delay loop architecture achieves two important functions. First it projects the input signal into a high-dimensional (HD) space with an up-sampling followed by an affine mapping with random bias and weight (W). As with other RC architectures, this HD projection function can separate entangled features in the low-dimensional subspace onto subspaces in the HD by hyperplanes, making it easier for classification. Second, each of STARE’s multiple loops operates under a different time-scale. Due to parallel processing of the loops, the state from the loops are projected into different hyper-spaces. The overall effect is equivalent to partitioning a larger hyperspace into L sub-hyperspaces, where each would focus on a different context. 

As shown in Figure 2, STARE has 3 main sections: 1) Input mapping and hyper-dimension (HD) projection; 2) Scaled-time attention reservoir; and 3) Linear regression readout. The system operates in two phases of a) training and b) running (i.e. form predict/classify). We first describe these components in detail, followed with examples of how they operate in both training and execution phase. 

As shown in Figure 2, the STARE architecture consists of two parallel delay-loop reservoirs, each with N delay elements. Each loop is assigned a leaky parameter, which controls the contribution of new incoming data towards the delay loop state. The loop operation is as follows:  
\begin{align} 
  x_i(n+1) = (1 - \alpha_i) * x_i(n) + a_i * \tanh( (w_{in1} + w_{in2}*u_h(n) ) + x_i(n-N) ), \quad i=1,2,\ldots,L 
\end{align}

where $i=1,2, \ldots , L$ corresponds to the $i$-th loop. $N$ is the total delay of the loop,
$ \alpha \in [0,1]$ is the leaky factor, $w_{in1},  w_{in2}$,  are the random bias and
input mask respectively,  is the up-sampled input. The input time series is first
upsampled and each sample then multiplied by a random weight $w_{in2}$ to obtain .
Thus, the time scale of $n$ is running at a rate $N$ times that of $t$. This
upsampling operation projects the input signal into the hyper-dimensional space.
The reservoir consists of $L$ parallel delay loops, which is described by Eq. (1).
Each of the $L$ parallel paths in STARE has a separate time scale controlled by the 
leaky parameter $\alpha_i$'s. We first describe the simplest configuration where there are 2
parallel loops, as many of the important features of the architecture can be explained.
We will later generalize to 3 or more loops. 

In a 2-loop configuration, Loop 1 (shown as shaded area in Figure 2), can be considered as the \emph{Short-term loop}, which
has a large leaky parameter ($\alpha_1$). From eq(1), it can be seen that the term $(1- \alpha_1)$ retains a small part of $x_1(n)$,
which will be used to compute the next state. The short loop thus remembers little of the past information. Conversely, Loop 2, which is designated as the Long-term loop, has a large leaky parameter ($\alpha_2$),
thus corresponds to having a long-term memory. 

Extracting temporal features by the two parallel loops also allows the long-term loop to extract “contextual” information from the input signal, 
and the short-term loop to track near-term fast changes (non-contextual). As an example, 
if context information is hidden in lower frequency component of a time series, the long-term 
loop extracts this low frequency component. As shown in Figure 2, the outputs of the serial states
 of two loops are $x_1(n)$ and $x_2(n)$, which are concatenated to form a $2N$-bit vector ($X$) and fed into the linear regression unit for training.

\subsection{Delay-loop Tree-Structure }

The idea of Delay-loop Tree (DL-Tree) configuration is motivated by comparing the different approaches of the shallow architecture of STARE and the Deep Neural Networks such as CNN. The key concept of the STARE approach is that by projecting the signal into the hyper dimensional space, it can be readily separated by a linear regressor as the signals that are tangled in the original signal space become separated in the hyper-dimensional space. However, a direct way to obtain the full benefit of hyper dimensionality with the SLR configuration leads to higher computational complexity. We shall show in the next section that for classification of RF modulations (dataset from DeepSig 
\cite{o2018over5}), such dimensionality requirements can be as high as 18,000. To implement such large dimensionality can defeat the original purpose of using a shallow neural network approach, as complexity rises linearly with dimensionality but accuracy improvement approaches a bound asymptotically.
\begin{figure}[h!]
  \centering
  \includegraphics[width=15cm,height=6cm]{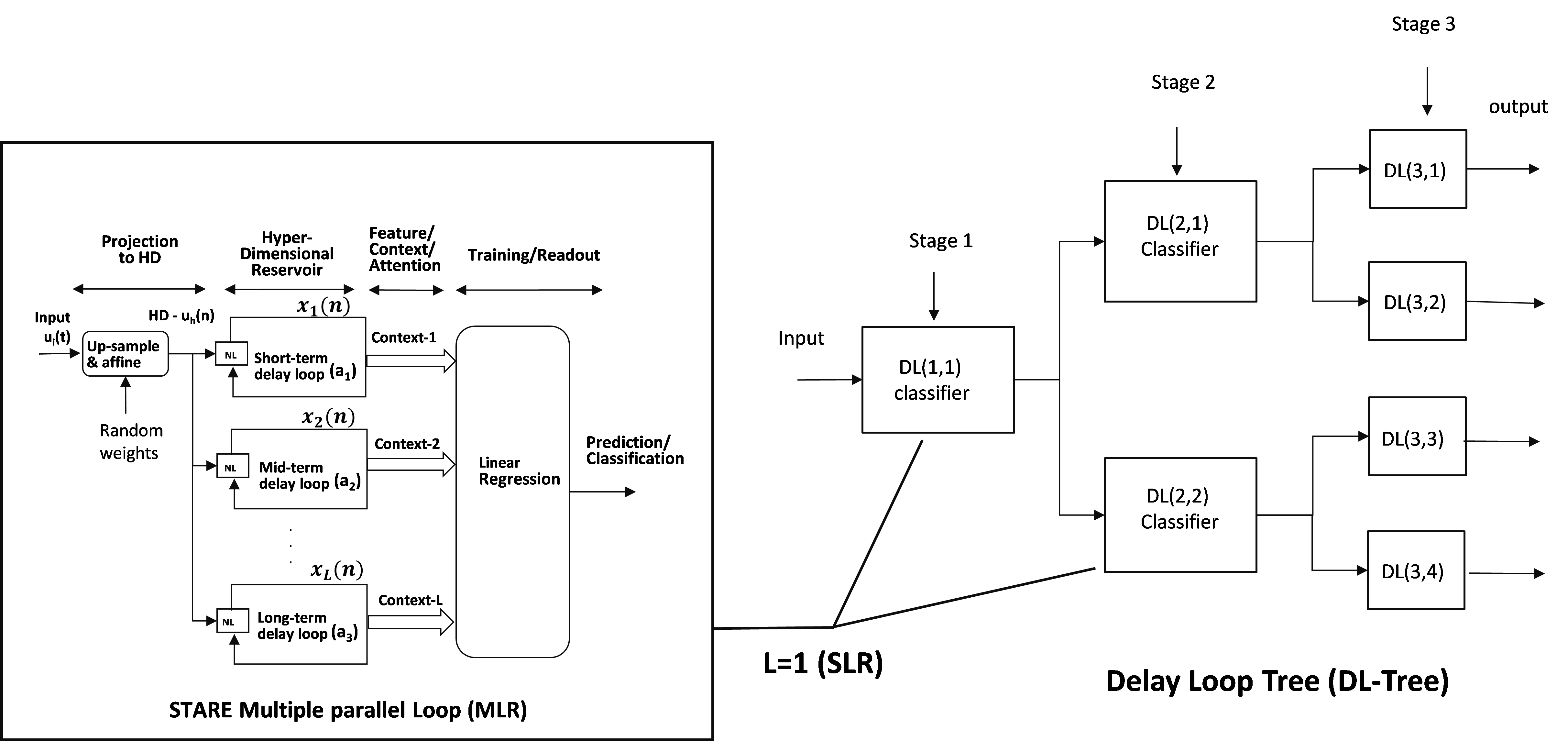}
  \captionof{figure}{STARE Architecture}
  \label{fig:fig3}
\end{figure}

The innovative approach of this work is to modify the 
STARE DL into a hierarchical tree structure composed of modular DLs. 
The main idea of DL-Tree is to partition the input into groups at each 
junction of the tree. At each level, the input signal is classified into 
subgroups. For example, in the first level, as shown in the right side of 
Figure 3, DL(1,1) is a SLR that classifies the DeepSig dataset 
into two groups of size G21 and G22 classes. The elements in each of
 the subgroup (G21, G22) is pre-determined by a heuristic algorithm 
 as described below. In stage 2, two classifiers are used to further 
 classify into 4 sub-groups (G31, G32, G33, G34), and so on. In general, 
 the DL-Tree may terminate at any level depending on the outcome of the
  heuristic algorithm. For example, we found empirically 
that for the 11-class DeepSig dataset, two stages would be optimum.
\begin{figure}[h!]
  \centering
  \includegraphics[trim={0.5cm 1cm 0 0.8cm},clip,
  width=15cm,height=6cm]{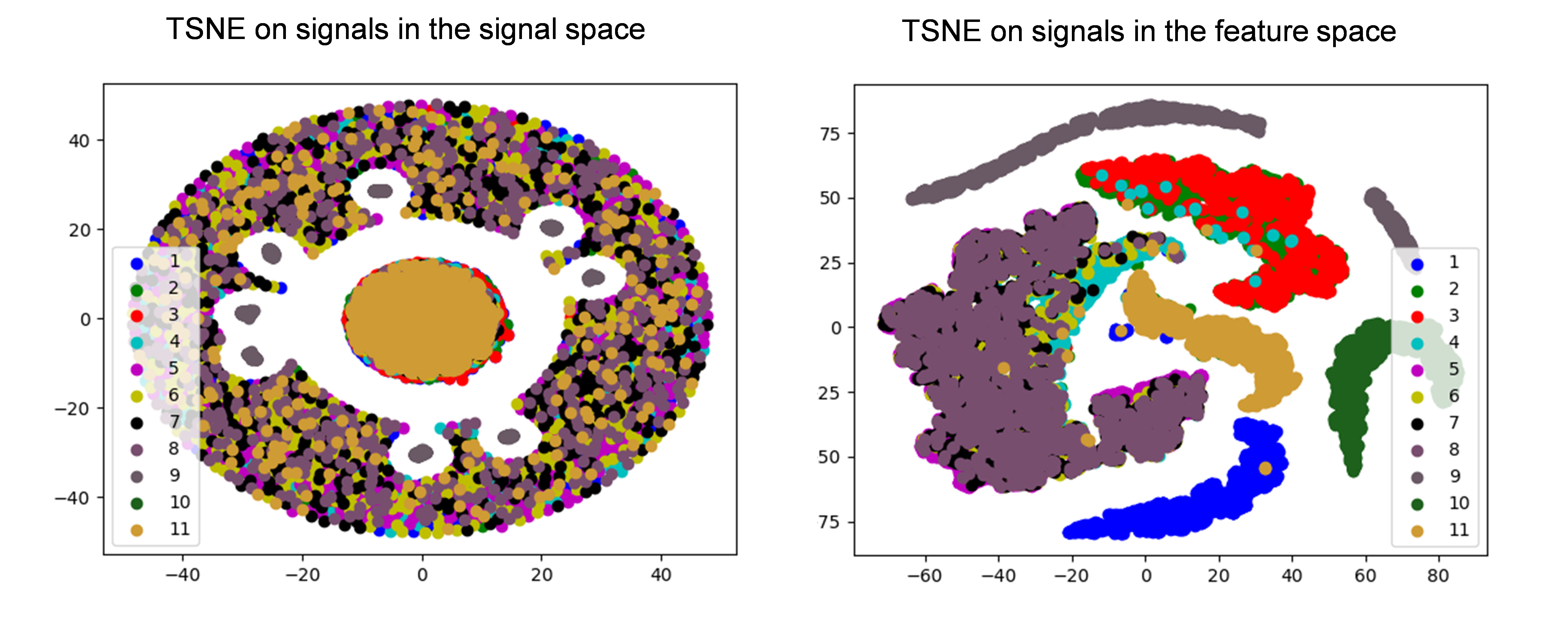}
  \captionof{figure}{Left: Feature in signal space. Right: Projected signal in DL feature space.}
  \label{fig:fig4}
\end{figure}

\subsubsection{Delay-loop Tree Heuristic Algorithm}

As shown in Figure 4, the TSNE (T-distributed Stochastic Neighbor Embedding \cite{van2008visualizing7})
 mapping of signals can identify signal classes with overlapping densities 
 and signals that seem to be well separated. Figure 4a shows the TSNE plot of the signal space of the RF modulated signals from the DeepSig dataset, consists of: OOK, 4ASK, BPSK, QPSK, 8PSK, 16QAM, AM-SSB-SC, AM-DSB-SC, FM, GMSK, OQPSK, where we see that the signal classes have overlapping
densities. As such, we can see that the signal classes are distributed in such a way that signals often have many neighbors that are in different classes. A consequence of this is that simple classification techniques, such as regression, do not work well on this dataset. In the TSNE plot of the projected feature space (Figure 4b), there still exist classes with overlapping densities, but there is also the addition of signal classes that are well separated from other signal classes. Empirically, we see that we are able to classify the well separated classes (i.e. classes 1 and 2) extremely accurately using the original STARE algorithm. However, the classes with overlapping densities (i.e. classes 6 and 7) in particular tend to be ones that are difficult to identify and are where the vast majority of misclassifications lie in a SLR. We introduce multistage classification strategies as modified STARE algorithms and see that they will achieve accuracies similar to a very high-dimensional SLR at a much lower dimensionality. Notably, the components of the multistage classification are much the same as the original STARE implementation. We only require performing linear regression on slightly different targets and chaining implementations of SLR in series.
A consequence of this is that simple classification techniques such as regression 
do not work well on this dataset. In the TSNE plot of the feature space (Figure 4b),
there still exist classes with overlapping densities, but there is also the addition
of signal classes that are well separated from other signal classes. Empirically,
we see that we are able to classify the well separated classes (i.e. classes 1 and 2)
extremely accurately using the original STARE algorithm. However, the classes with
overlapping densities (i.e. classes 6 and 7) in particular tend to be ones that
are difficult to identify and are where the vast majority of misclassifications
lie in a SLR. We introduce multistage classification strategies as modified STARE algorithms and see that they will achieve accuracies similar to a very high-dimensional SLR at a much lower dimensionality. Notably, the components of the multistage classification are much the same as the original STARE implementation. We only require performing linear regression on slightly different targets and chaining implementations of SLR in series. 

In the DL-Tree configuration, we describe the case of two stage classification which will hold in generality for a higher number of stages by recursion. In two stage DL-Tree classification, we consider partitioning the possible classes into groups. We can then run two levels of classification – first to identify which partition group of classes the signal belongs to and second to identify which class the signal belongs to conditioned on the partition group has been placed in. The motivation behind this is to hopefully separate out “easy” classes from the problem and be left with only the “hard” classes left to classify. Then the predictive power of the linear readout after the HD projection can be maximized in this focused classification problem. Notably, we do not need the additional predictive power on the “easy” classes as a lower dimensional SLR already does well enough on that subset.

The first stage performs SLR algorithm with a different regression target. In the linear regression, we have defined the new target as the logical OR of the columns of classes in the same group of the partition. 
Letting ; denote concatenations, $P_k$ denote groups of the partition, and $Y_i$ denote columns of the original target $Y$,
 we have that the new target is $[V_{i \in P_1}Y_i; \ldots; V_{i \in P_k}Y_i]$. The second stage again proceeds much like a smaller SLR over the classes in each group of the partition.

We use various heuristics to determine a good partition to use in the
multistage classification. Recalling the motivating idea, we would like to
see that the first stage classifier groups classes such that (1) the accuracy 
of the first stage is high (2) confounded and difficult classes are grouped together 
(3) groups are of nontrivial size. This leads to a natural heuristic in terms of
the confusion matrix of the first stage classifier. Let $\alpha$ denote a 
hyperparameter tradeoff factor, $C­_{ij}$ denote entries of the confusion matrix,
and $P_k$ denote the groups resulting from choice of the partition $P$. Then we 
have the natural optimization problem over the candidate partitions
  \begin{align} 
      \min_p \sum_{k=1}^n \left\{ \alpha \sum_{i \in P_{k} , j \notin P_k} [C_{ij} +C_{ji} ] - (1-\alpha) 
      \frac{\sum_{i,j \in P_k}[C_{ij}+C_{ji}] - \sum_{i \in P_k}C_{ii} }{|P_k|} \right\} 
  \end{align}
When we look at the result of this optimization problem, we see that this picks classes that have the highest prediction entropy in SLR. In a similar sense, we can think of this heuristic as the simultaneous maximization of an analog of entropy in the partition groups and minimization of cross entropy of the first stage classification. In a more abstract sense, we can also view this as finding a permutation of the confusion matrix with dense block diagonal. 
\begin{figure}[h!]
  \centering
  \includegraphics[trim={0.1cm 0 0 0.5cm},clip, 
  width=14cm,height=6cm]{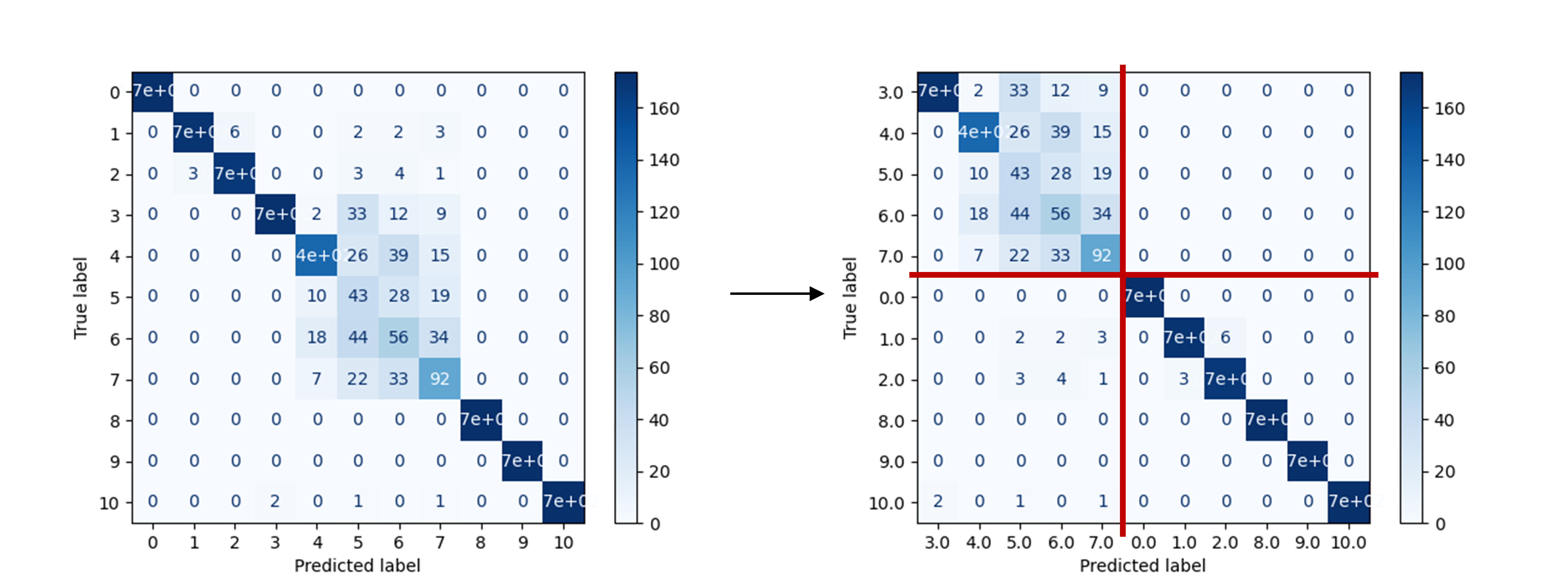}
  \captionof{figure}{Illustration of Heuristic algorithm for DL-Tree partitioning}
  \label{fig:fig5}
\end{figure}

As seen in Figure 5, this heuristic suggest grouping together classes 3, 4, 5, 6, 7. When referenced with Figure 4, we see that this coincides with our belief of which classes are “hard” as also demonstrated by the density of the matrix block. We may further recurse on this group to get more focused classifiers but we found that there is no significant improvement over the two stage classification. 
\subsection{Non-MAC computation in STARE }

Neural network training and inference are typically performed by multiply and accumulate (MAC) resources found in CPUs and GPUs. The limited availability of these resources usually dictate performing these calculations serially resulting in longer computation times. The STARE architecture lends itself to requiring lower precision multiplications in its reservoir calculations allowing for a non-MAC implementation utilizing look-up tables. At resolutions below 12 bits, non-MAC implementations utilizing look-up tables result in very significant cost, power and resource savings as well as speed improvements. Figure 6 shows DeepSig classification performances error due to limited precision non-MAC calculations for 6 to 16 bits. Significant performance error is not seen until the resolution drops below 8 bits and can be reduced further by increasing the reservoir size. Additional advantages of the non-MAC implementation are described in Section 4.

\begin{figure}[h!]
  \centering
  \includegraphics[width=7cm, height=6cm]{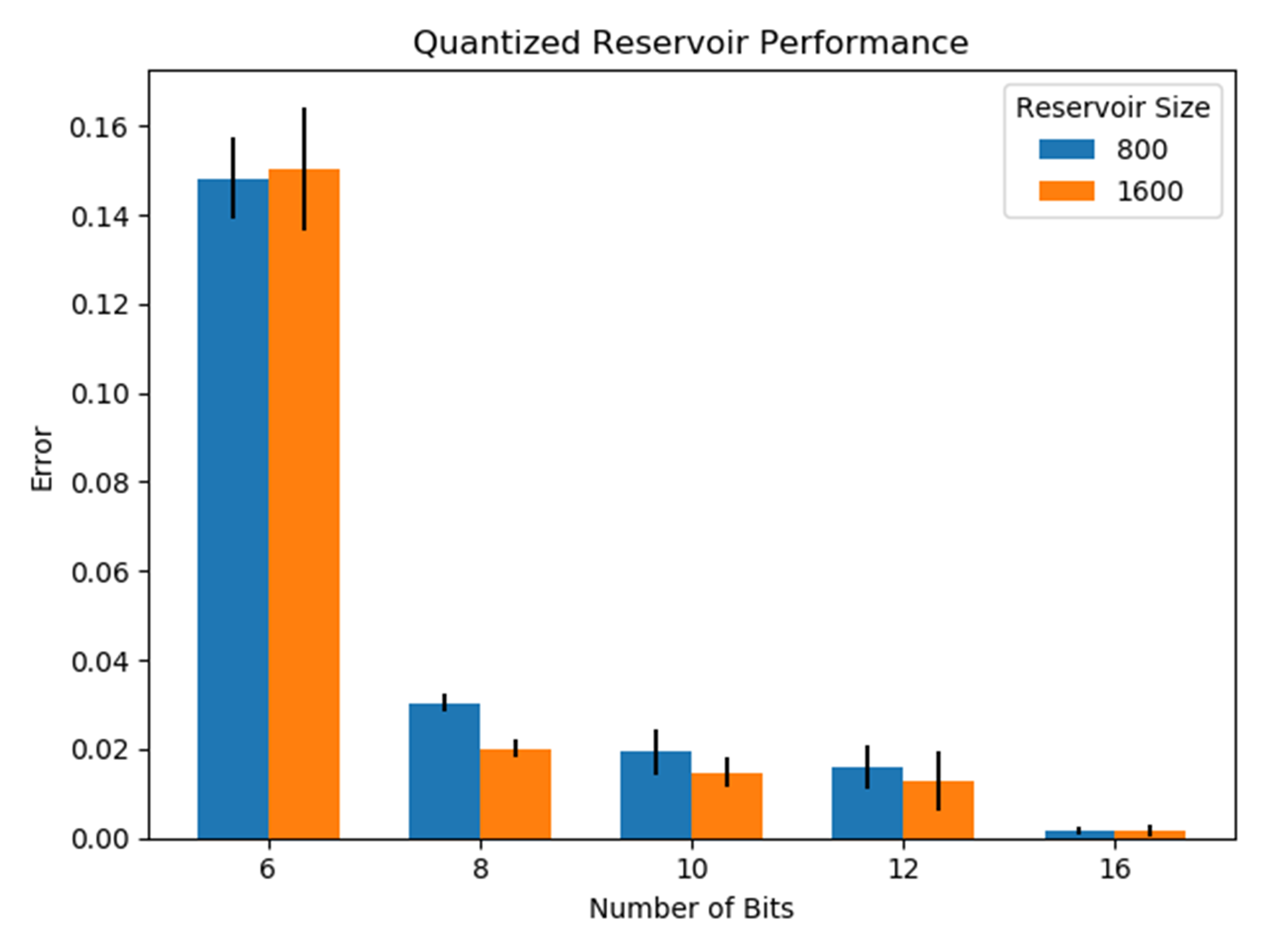}
  \captionof{figure}{Performance of Quantization in STARE}
  \label{fig:fig6}
\end{figure}

\section{STARE Simulation and Results }

We simulated STARE for 3 applications and demonstrated the result and when appropriate compared to the SoA performance.

\subsection{Time Series Prediction}
\subsubsection{Modulated Mackey Glass}

\begin{align}
  \frac{dx}{dt} = \beta \frac{x_r}{1+x^n_r} - \gamma x \quad \gamma, \beta, n > 0 
\end{align}

Chaotic signals such as the Mackey Glass (Eq. 3) are a common bench mark test for prediction. A clip of the original Mackey Glass signal is shown in Figure 7(a). In the following we will first show that most NNs perform well in the prediction of the Mackey Glass signal. We next modulate the Mackey Glass with a cosine signal (period of 500) which we will refer to as the modulated Mackey Glass signal (mMG). This low-frequency modulated chaotic signal exhibits a non-stationary property that is usually challenging for prediction algorithms. We will show that for both LSTM and single delay-loop reservoir, predicting the original Mackey Glass signal is quite straightforward (with LSTM or single delay-loop approaches), but both fail to predict the modulated MG. In the case of a single-loop reservoir, the training did not converge at all. For LSTM, it converges but with large MSE. However, STARE was able to predict the mNG with low MSE.

\begin{figure}[h!]
  \centering
  \includegraphics[
  width=18cm,height=6cm]{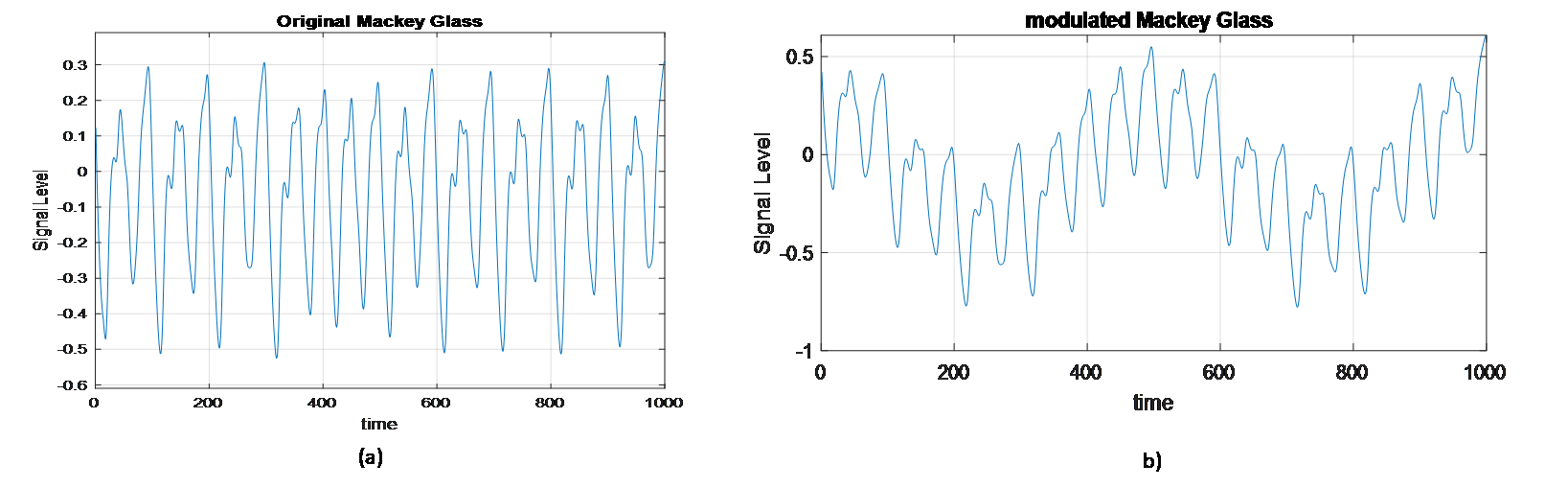}
  \captionof{figure}{Mackey Glass signals. a) Original. b) Cosine-modulated with a period of 500}
  \label{fig:fig7}
\end{figure}

We apply STARE algorithm using 2-loops configuration. The leaky factor for the long loop 
is $\alpha_1=0.97$ and that for the short loop is $\alpha_2=0.2$. We use 3000 samples for training
and predict the following 1000 samples. The prediction result, as shown in Figure 7(a)
is very good and the Mean Square Error (MSE) is found to be 0.03. It is seen
that the initial 500 samples of prediction is extremely accuracy. Prediction accuracy degrades over time,
but the prediction trend (context) is still very good. This observation is further verified in Figure 7(b), 
where the internal state of the 2 loops are plotted.
 It is clear that the state of the short loop (red) changes with the same pace as the fast component of the signal and that the state of the long loop (blue) follows the frequency of the low-frequency component. 

\begin{figure}[h!]
  \centering
  \includegraphics[trim={0 0 0 0},clip,
  width=18cm,height=6cm]{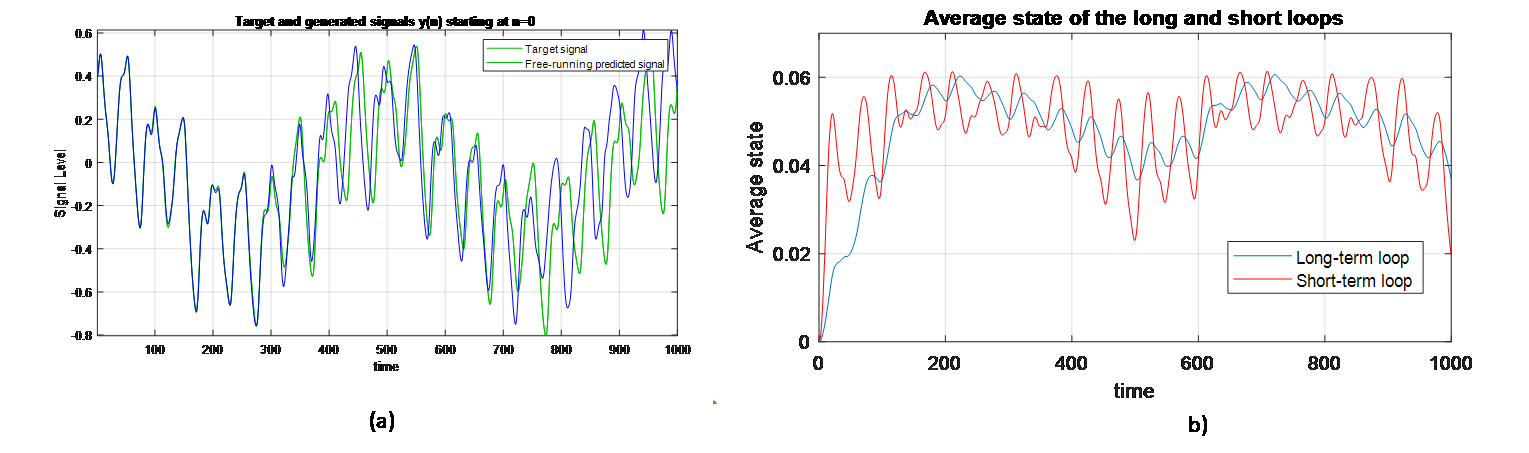} %
  \captionof{figure}{Prediction of mMG using dual-loop STARE. a) Prediction vs Original. b) State of long vs short loop}
  \label{fig:fig8}
\end{figure}

\subsection{Comparing with LSTM}

We trained our autoregressive LSTMs on sequences of 1000 samples for each oscillator type. We then seeded each model with a previously unseen sequence of length 1000 and ran the model in open-ended mode to generate 1000 new samples. We evaluated the model by comparing the model’s predicted sequence of samples with the actual subsequent samples, both visually and with MSE. We also trained a third LSTM model on sequences of length 2000 from the modulated oscillator to see if this helped identify the modulation.

The first LSTM is able to mirror the unmodulated Mackey-Glass oscillator very well out to about 300-400 samples. The predictions diverge somewhat after that point, which is not surprising because Mackey-Glass is chaotic, and small errors tend to get amplified over time. However, the LSTM still captures the general shape of the signal throughout the generated samples and maintains a low MSE.

Signal modulation adds meaningful complexity to the classification. While the LSTM again picks up the general shape of the Mackey-Glass signal, it almost completely ignores the modulation, which is reflected strongly in the MSE. Training on longer sequences was not any better. While there is some low frequency signal present in the predictions, it does not pick up on the periodicity of this signal and does worse than the short sequence LSTM with respect to MSE.

\subsection{DeepSig Modulation classification (Richard)}

We explored the space of delay loop architectures in a wide variety of dimensions with a goal of matching the performance of ResNets on the 11 class 
DeepSig18 data \cite{deepsig6}. For this comparison, the delay loops in this section do not use our non-MAC quantization scheme; rather they use single-precision floating point (32 bits) computations. The DeepSig18 dataset has a wide range of SNRs, spanning from -20 dB to nearly noise-free 30 dB data. We focus our evaluations on an intermediate SNR level to most effectively differentiate between different algorithms. Unless otherwise stated, the results in this section are on 6 dB data. For comparison, the ResNet reported an accuracy of 
96\%, which is representative of the state of the art.
\begin{figure}[h!]
  \centering
  \includegraphics[width=12cm,height=8cm]{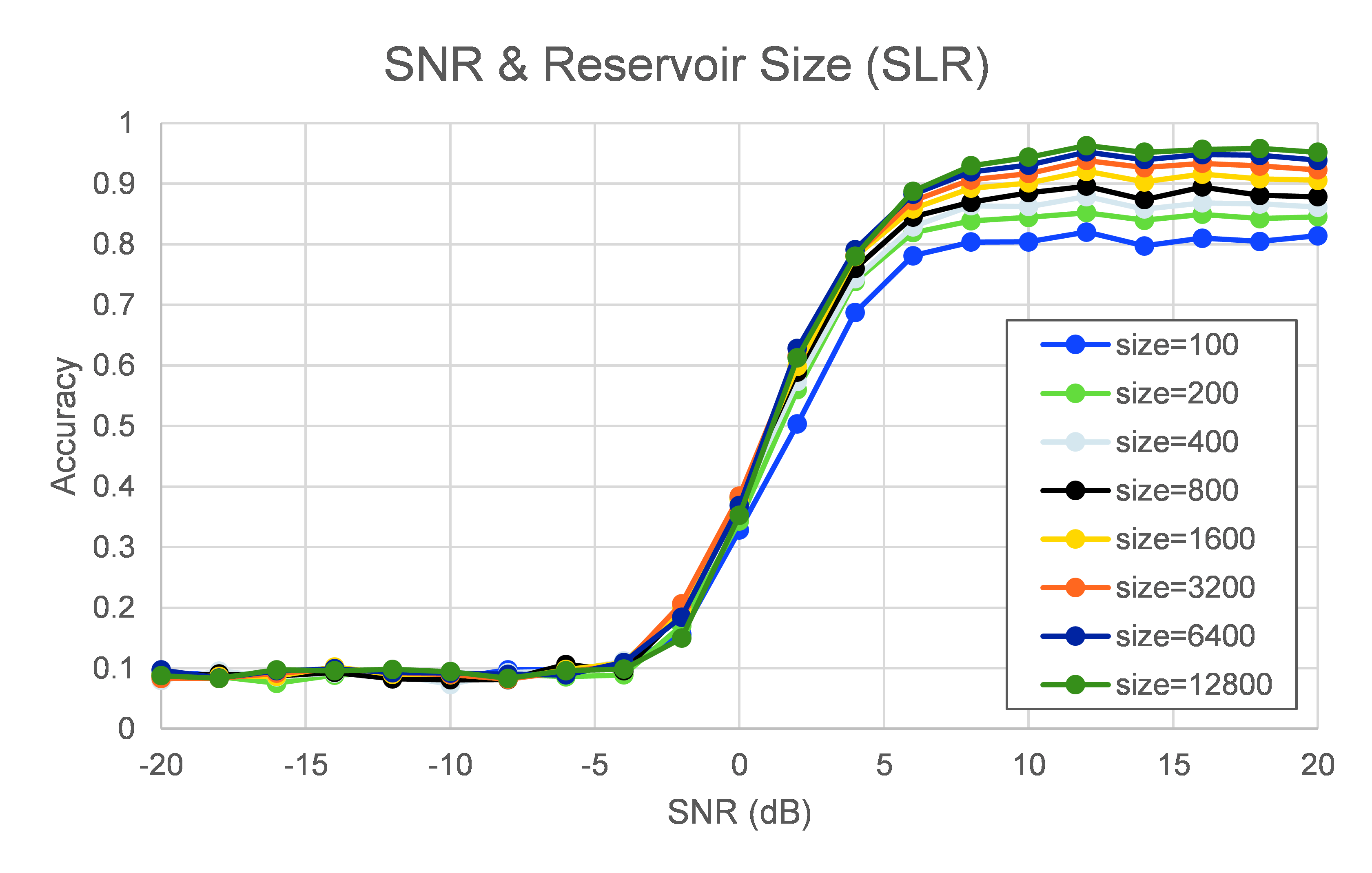}
  \captionof{figure}{Performance of STARE on DeepSig Classification vs SNR and different reservoir sizes}
  \label{fig:fig9}
\end{figure}

\subsubsection{Performance with Noise }

Figure 9 shows the STARE SLR performance with respect to different SNR levels and different DLR sizes. The model is trained with noisy samples, so this is expected behavior. It can be seen that the delay loop architecture decays gracefully in the presence of noise. Overall the accuracy improves with higher dimensionality. Compared to the SoA ResNet implementation, the STARE performance is within 
6\% for dominated by the noise, so any model, including STARE and our baseline ResNet, struggles to do much better than random guessing.

 \subsubsection{ Comparing SLR vs Tree-DL on DeepSig Classification}

 We tested both SLR and DL-Tree configurations. The result is shown in Figure 10, in which the blue curve is obtained by SLR and the orange curve from the DL-Tree classifier. While both configurations approach about 
 90\% accuracy, the DL-Tree dramatically improves performance on small sized delay loops (1600-3200 nodes). This allows us to achieve good performance with much smaller delay loops.

 \begin{figure}[h!]
  \centering
  \includegraphics[width=10cm,height=6cm]{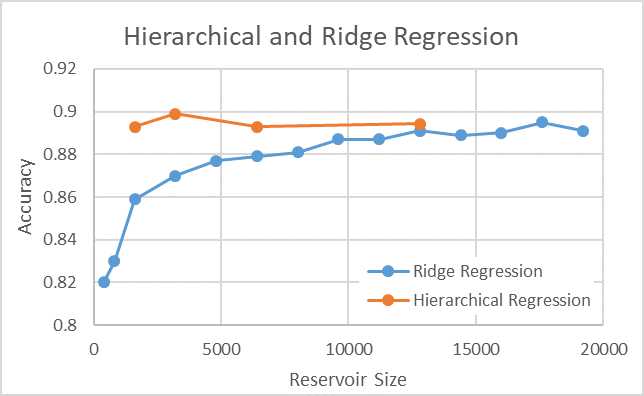}
  \captionof{figure}{Performance Improvement with hierarchical DL-Tree algorithm}
  \label{fig:fig10}
\end{figure}

\subsection{Counter-UAS Detection and Identification}
\subsubsection{Counter UAS Application}

\begin{wrapfigure}{L}{0.5\textwidth} 
  \centering
  \includegraphics[width=6cm,height=6cm]{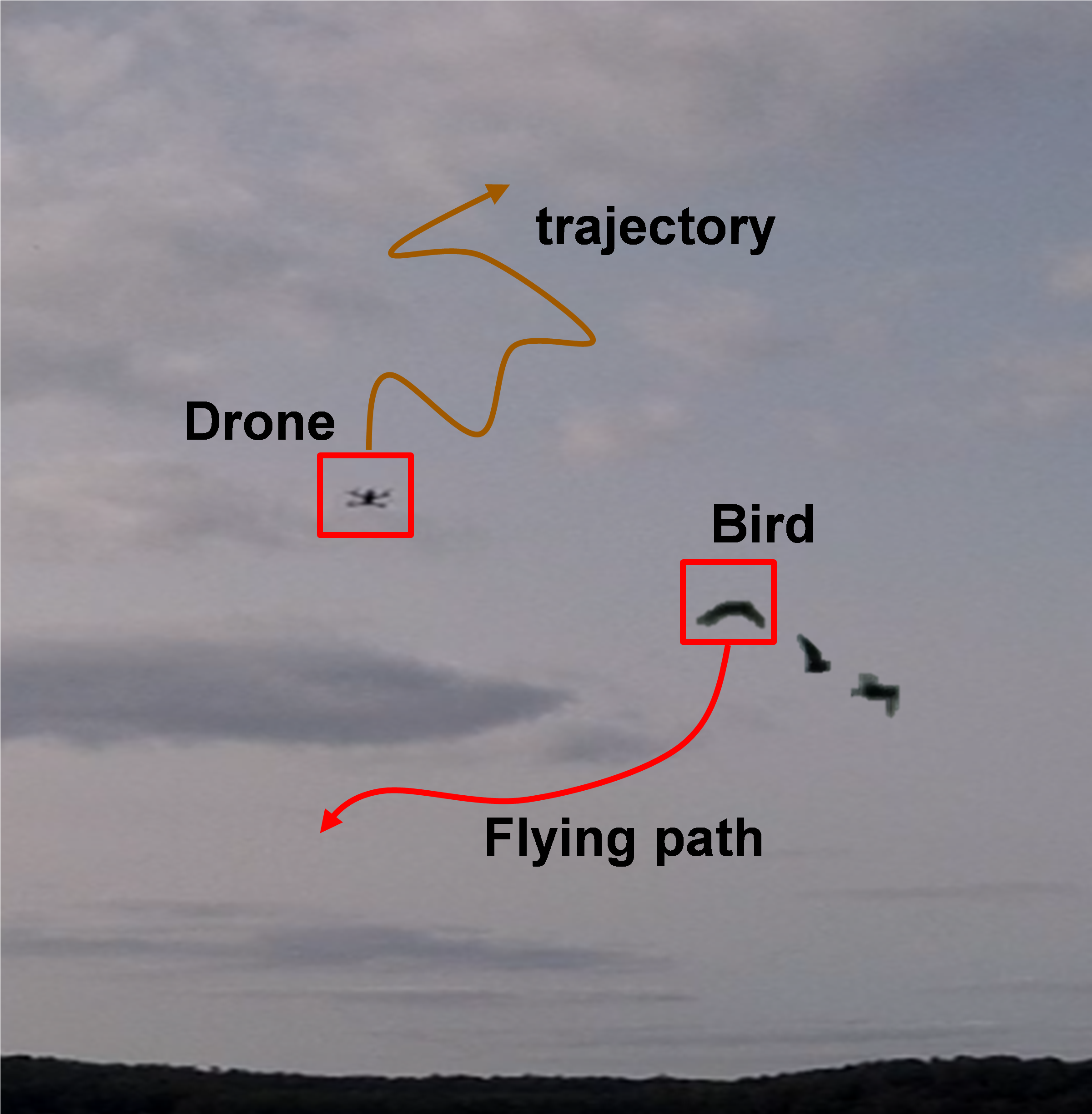}
  \captionof{figure}{Counter-UAS Detection}
  \label{fig:fig11}
\end{wrapfigure}

In the counter UAS detection and identification application as illustrated in Figure 11, it is important to be able to detect and classify flying objects at a large standoff distance and alert the defense system. Radar can be used to detect foreign objects at large distance, but it is challenging for radar to detect small and slow-flying objects and accurately distinguish drones from birds \cite{rahman2018radar9}. EO/IR camera, on the other hand, can only operate in ranges ~500 meters, which is not sufficient for counter-UAS applications. RF sensors rely on detection of the RF radiation from drones and can be shown to work well within 1km. Audio sensors work well at night time, but are effective only within 50 meters.

In this section, STARE dual-loop is applied to bird/drone detection and identification. The key idea is visual frames and on-screen movement obtained from birds and drones exhibit sufficient difference for the classification task. We utilize imagery and movement trajectory information from video cameras to address the issue of long standoff detection/classification, and demonstrate the application’s robustness to data constraints and visual input noise through combined information from dual channels. See Data Format section 3.3.3.3 for image processing and trajectory extraction details and Classification Performance section 3.3.4 for performance comparison between visual and trajectory inputs.

Following the architecture diagram below, we incorporate both visual (Input 1) and movement trajectory (Input 2) information to classify the observed object. In the Results section, we show the utilization of both visual and trajectory data outperforms classification by visual or trajectory data alone.

\subsubsection{Dual-loop Architecture for Counter-UAS Detection}
\begin{figure}[h!]
  \centering
  \includegraphics[width=14cm,height=5.5cm]{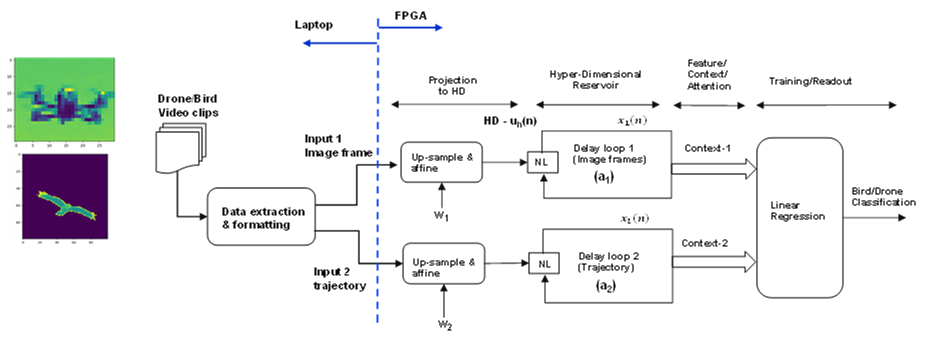}
  \captionof{figure}{DLR for Drone/Bird classification. Upper loop for image frame data, lower loop for trajectory data}
  \label{fig:fig12}
\end{figure}

Various approaches for counter-UAS were reported in the literature [7, 8]. However, most of these approaches rely on processing image frames from the footage. In our approach we take advantage of the dual-loop STARE architecture and extract features from both video frames and trajectory of the flying path. The dual-loop architecture is suitable for this new approach as different types of features are processed into separate DLR loops, and recombined after the relevant features are extracted. One of the research goals is to demonstrate the effect of combining both spatial and trajectory features for drone/bird classification. Thus we employ the STARE dual-loop architecture (Figure 12) to incorporate both visual (Input 1) and trajectory (Input 2) information to classify the observed object. In the following, we compare classification accuracy for visual data only, trajectory data only, and both visual and trajectory data.

This is motivated by the idea that classification evidence within visual and trajectory inputs are unique to the input, and thus, is best to be processed in separate DLR loops, so that each DLR can be configured differently to extract different features for classification. Specially, we choose a smaller leaky factor for the image loop and a larger one for the trajectory loop, as imagery features tend to change much faster than the features of the flying path.

\subsubsection{Data for Training and Testing}

Bird video data is a series of public, community collected footage featuring one or multiple birds in flight from Storyblocks.com. We used in total 5 different source videos containing 3 bird species: seagull, hawk, and flamingo. The bird class totals 4300 frames, with per-frame bird center coordinates, from which classification observations are created.

Because high quality video of drones in flight is not readily available, drone footage was created by Perspecta Labs. Footage was taken at a number of locations with various backgrounds including over water and residential backyard for a mostly static background. Four different types of drones were flown with diameters measuring 34 and 18 cm at their widest point including the propellers. Flights were performed at various distances from the camera in order to provide a variety of resolution scenarios. Additionally, both high speed and low speed maneuvers were performed to capture different flight characteristics.

\begin{figure}[h!]
  \centering
  \includegraphics[width=13cm,height=5cm]{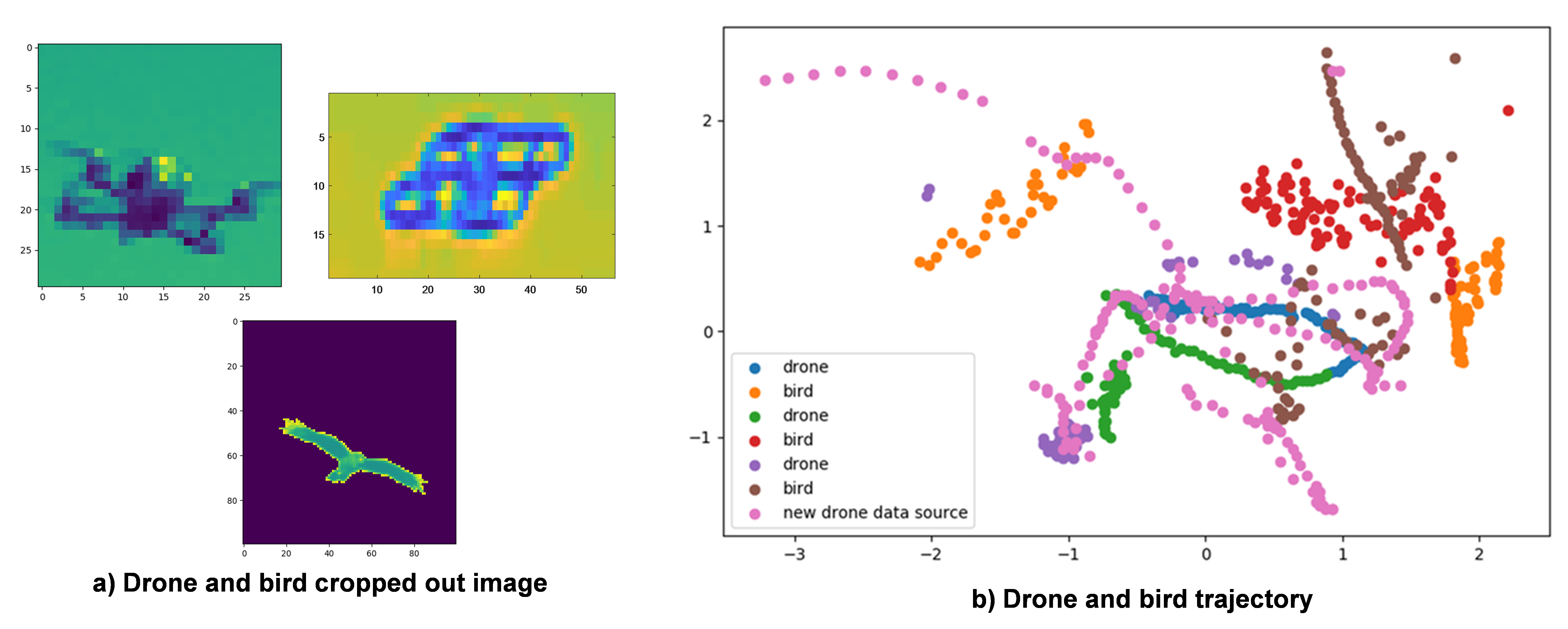}
  \captionof{figure}{a) Cropped drone/bird image data. b) Drone/bird trajectory data}
  \label{fig:fig13}
\end{figure}

\subsubsection{Data Format}

An observation ( a burst in the DeepSig context ) is an example in the video dataset to be classified which may take the form of an image sequence or a coordinate sequence denoting the flying object’s center (the object trajectory). Data processing below describes the preprocessing procedures, resulting in visual observations such as figure 13a, and trajectory observations such as 13b.

The image modality utilizes a hundred $30 \times 30$ frames by default. A key benefit of the STARE architecture is introducing additional information within the trajectory observation, which are by default 100 (x, y) coordinates denoting the target location within the screen.

\subsubsection{Classification Performance}
The STARE dual-loop algorithm is evaluated by varying data and model parameters including: reservoir size (dimensionality of the delay loop) and non-MAC fixed point quantization. In the following experiments, we consider two models: single delay loop ingesting image modality only (blue), and the dual loop ingesting image and trajectory modalities (red). All target images are 30 X 30 frames sampled at 10 frames per second.

\textbf{Reservoir Dimensionality and Noise} \\
For an effective Counter UAS application, weather often serves as obstacle towards successful target identification \cite{lau2015see8}. To demonstrate the efficacy of integrating trajectory information when images are degraded, weather situations such as snowstorm or heavy rain fall in the image modality are simulated by adding a gaussian noise mask to the target image.

By increasing the reservoir size, STARE performance in noisy visual environments improves. At 0 decibel signal-to-noise ratio (SNR db) in Figure 14(a), increasing reservoir size up to 1500 lead to noticeable gains in performance. For stronger noise, such as –4 SNR db in Figure 14(b), early performance improvements from increasing reservoir size become more substantial.

\begin{figure}[h!]
  \centering
  \begin{subfigure}{.5\textwidth}
    \centering
    \includegraphics[width=6cm,height=5cm]{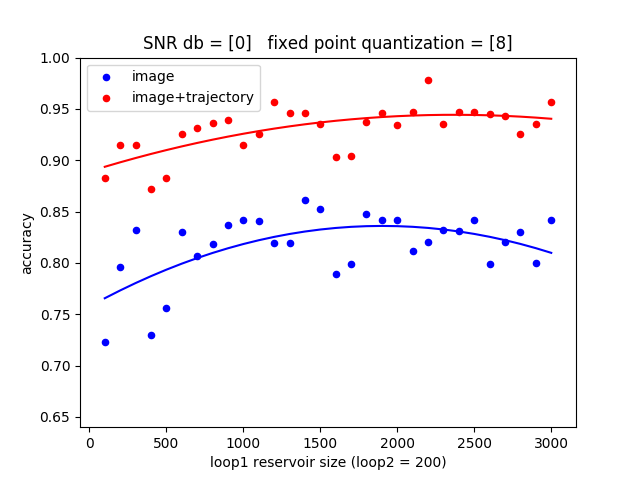}
    \label{fig:fig14a}
  \end{subfigure}%
  \begin{subfigure}{.5\textwidth}
    \centering
    \includegraphics[width=6cm,height=5cm]{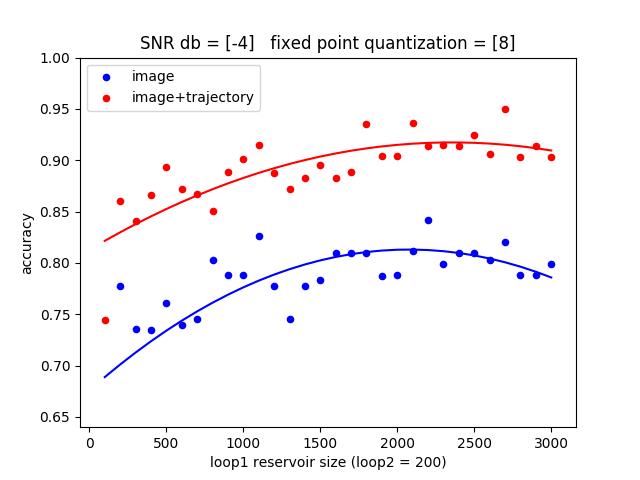}
    \label{fig:fig14b}
  \end{subfigure}
  \caption{Drone/bird classification performance as reservoir size increases. A) Image modality is corrupted by 0 SNR db, where power of gaussian noise equal to the power of the data. B) Image modality corrupted by -4 SNR db.}
\end{figure}

\textbf{Non-MAC Quantization} \\
Figure 15(a) displays performances for the typical 8-bit quantization case. The x-axis, from left to right, implies decreasing noise. A lower precision quantization, such as 4-bit in Figure 15(b), decreases the performance under all noise conditions by noticeable amount when compared to the 8-bit quantization.

\begin{figure}[h!]
  \centering
  \includegraphics[width=13cm,height=5.5cm]{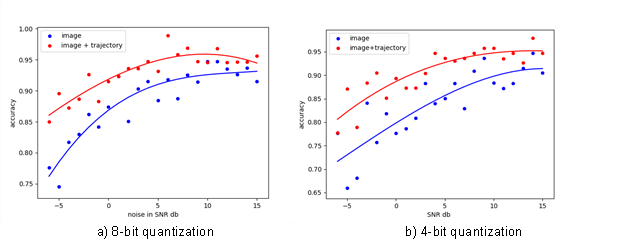}
  \captionof{figure}{Drone/bird classification performance vs. video at different SNR levels and two quantizations. A) 8-bit fixed point quantization and B) 4-bit fixed point quantization.}
  \label{fig:fig15}
\end{figure}

\section{Implementation of STARE}
\subsection{FPGA}

Modern field-programmable gate arrays (FPGA) contain large arrays of programmable logic blocks, which are typically in the form of small look up tables (LUTs) that can be programed and interconnected by loading a configuration file. Memory blocks, digital signal processing blocks, and clock generation/synchronization blocks are also included in the fabric of the gate array. This array is often surrounded by peripherals in the form of high speed data interfaces and an embedded control processor. Programing of the array is done through a hardware descriptive language like VHDL or Verilog.

\subsubsection{Serial FPGA Implementation of STARE Loop}
The equations that describe the STARE loop operation lend themselves to both a serial and a parallel implementation within the FPGA. The parallel implementation is of particular interest because the reduction in required resources due to a quantized non-MAC approach allow for significant increases in throughput relative to a serial implementation without unmanageable increases in resource utilization. Because the serial implementation requires few resources, it can be quickly implemented within an FPGA and is better suited to experimentation because of the short design cycle between code changes and the ability to test results.

Figure 16 shows a block diagram of the serial implantation of a STARE loop as described by the following loop equation.
\begin{align} 
x_i(n+1) &= (1-\alpha_i)*x_i(n-(N-1))+ \\
 \alpha_i &\tanh( (B_{in}(n\%N) + w_{in}(n\%N) * u_h(n)) + x_t(n) ), \quad i=1,2,\ldots,L,n\%N =0 \nonumber \\
        &=  (1-\alpha_i)*x_i(n-(N-1))+ \nonumber \\ 
 \alpha_i & \tanh((B_{in}(n\%N) + w_{in}(n\%N) * u_h(n)) + x_t(n-N) ), \quad i=1,2,\ldots,L,n\%N \neq 0 \nonumber
\end{align}

where $i=1,2$ corresponds to the long-term and short-term loop respectively. $N$ is the total delay 
of the loop, $\alpha_i \in [0, 1]$ is the leaky factor, $B_{in}$ and $w_{in}$ are the length $N$ random bias and input masks.
$u_k(t)$ is the input time series and $u_h(n)$ is the up-sampled input where $u_h(n) = u_k(\lfloor \frac{n}{N} \rfloor )$. 
Every $N$ samples, an output vector $\protect\ora{X_i}(t)$ is formed from the last $N$ values of $x_i$.

\begin{figure}[h!]
  \centering
  \includegraphics[width=14cm,height=5cm]{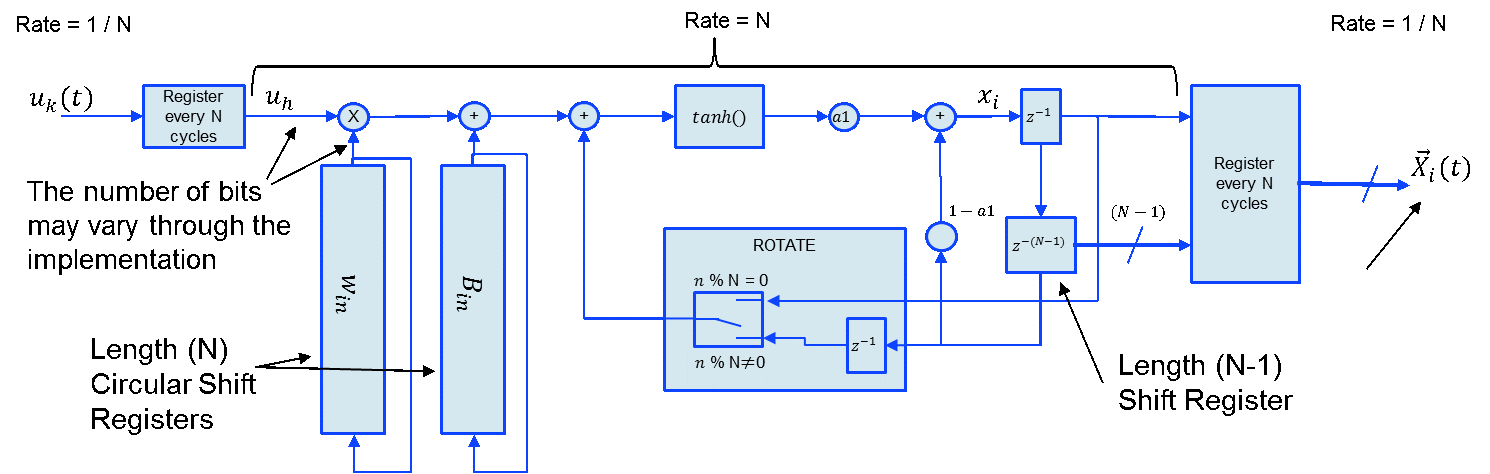}
  \captionof{figure}{Block diagram of a proposed serial implementation of a STARE loop.}
  \label{fig:fig16}
\end{figure}

As can be seen from Figure 17, each loop can be efficiently implemented in a serial fashion
 within an FPGA using a single two input multiplier, two constant coefficient multipliers
($\alpha_i$ and $(1-\alpha_i)$), three adders, and a single implementation of the $\tanh$ function,
 along with register storage and some control logic.  Although this structure is efficient
  from a resource utilization perspective, it requires at least $N$ clock cycles to
   process each element of the input time series $u_k(t)$, thus reducing the maximum possible
    throughput of the loop by a factor of $N$ relative to the parallel implementation
     described in the next section.  

Note that, even with the serial implementation, throughput gains can be achieved by including multiple copies of the loops in the FPGA fabric, thus allowing more than one sample to be trained or classified simultaneously.

\subsubsection{Parallel FPGA Implementation of STARE loop}

The loop described in Eq. (4) can equivalently be calculated in a parallel fashion using
 the following equation, which produces the same set of output vectors $\protect\ora{X_i}$ as (4).
\begin{align}
  \protect\ora{X_i}(t+1) = (1-\alpha_i)\protect\ora{X_i}(t) + \alpha_i \tanh( ROTATE(\protect\ora{X_i}(t)) + \protect\ora{W_{in}}*u_k(t+1) + \protect\ora{B_{in}} )  
\end{align}

Figure 17 illustrates a parallel form for computing the bold portion of the equation 
\[
  (1-\alpha_i)\protect\ora{X_i}(t) + \alpha_i \tanh( ROTATE(\protect\ora{X_i}(t))
   + \pmb{ \protect\ora{W_{in}}*u_k(t+1) + \protect\ora{B_{in}}  } )
\]
\begin{figure}[h!]
  \centering
  \includegraphics[width=9cm,height=5cm]{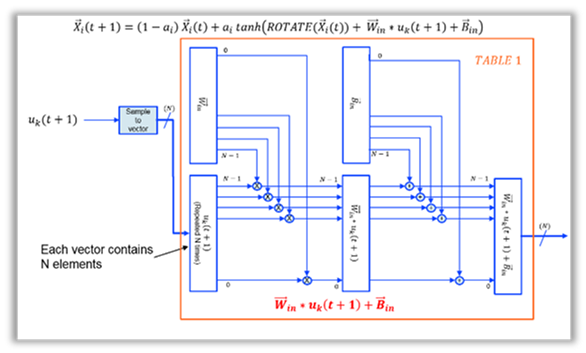}
  \captionof{figure}{parallel form for computing $\pmb{ W_i * u_k(t+1) + 
  \protect\ora{ B_{in}} } $.}
  \label{fig:fig17}
\end{figure}

As Figure 17 shows, most operations can be performed in parallel,
 at the cost of requiring additional computing resources within the FPGA 
 relative to a serial implementation.  For example, $N$ multipliers and 
 $N$ adders are required to compute this portion of the equation.  Importantly,
  because $\protect\ora{W_{in}}$ is a constant during the run of the reservoir,
   the multiplications in this section can be performed with constant coefficient
    multipliers, which can be efficiently implemented as look up tables when the
     number of bits in the multiplier is low.

\begin{figure}[h!]
  \centering
  \includegraphics[width=10cm,height=5cm]{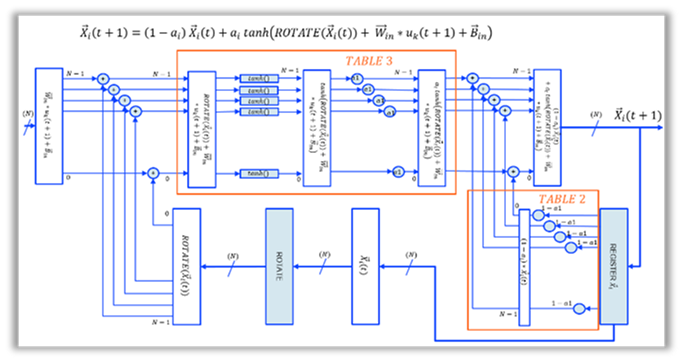}
  \captionof{figure}{Parallel implementation of highlighted part of 
  $(1-\alpha_i)\protect\ora{X_i}(t) + \alpha_i \tanh( ROTATE(\protect\ora{X_i}(t)) + 
   \pmb{ \protect\ora{W_{in}}*u_k(t+1) + \protect\ora{B_{in}}  } ) $ }
  \label{fig:fig18}
\end{figure}

As shown, an additional $2N$ constant coefficient multipliers are required along with
 an additional $2N$ adders.  The $ROTATE$ function in the equation is merely an inexpensive routing
  operation within the FPGA fabric.  Notably, $N \tanh()$ functions are required, which
   would be extremely expensive in terms of resources if lower bit quantization’s did not allow for an efficient look up table based approach to computing this function.  Accordingly, the parallel implementation may not be well suited when using floating point arithmetic or integers with a large number of bits.

   The advantage of the parallel implementation is that, ignoring device speed, it is possible
 to compute the next output vector $\ora{X_i}$ in a single clock cycle, making the throughput of the
  parallel implementation at least $N$ times faster than that of the serial version.  Hybrid approaches are
   also possible to more finely tune the tradeoff between resource utilization and throughput.

\subsubsection{FPGA Resource and Throughput}
A Xilinx® Zynq UltraScale+ RFSoC ZCU111 development board was used to implement a parallel version of the STARE algorithm. This platform was chosen because it combines a large FPGA fabric with a tightly integrated processing system. We first describe a single-loop DLR with a 512-bit reservoir (dimensionality N) that was implemented in the FPGA fabric using an 8-bit fixed-point non-MAC approach. Three lookup tables were used in each of the 512 reservoir cells. Figures 17 and 18 show what parts of the parallel algorithm are contained in the three tables.

A client PC was used to transmit blocks of data over an Ethernet interface to an ARM processing core that resides in the FPGA. This data is filled into input block rams in the FPGA fabric that are connected to the reservoirs. The output of the reservoir is captured in output block ram and then sent back over the Ethernet interface to the Client PC. In the case of a dual loop DLR, there are two copies of the reservoir in the fabric as shown in Figure 19.
\begin{figure}[h!]
  \centering
  \includegraphics[width=12cm,height=5cm]{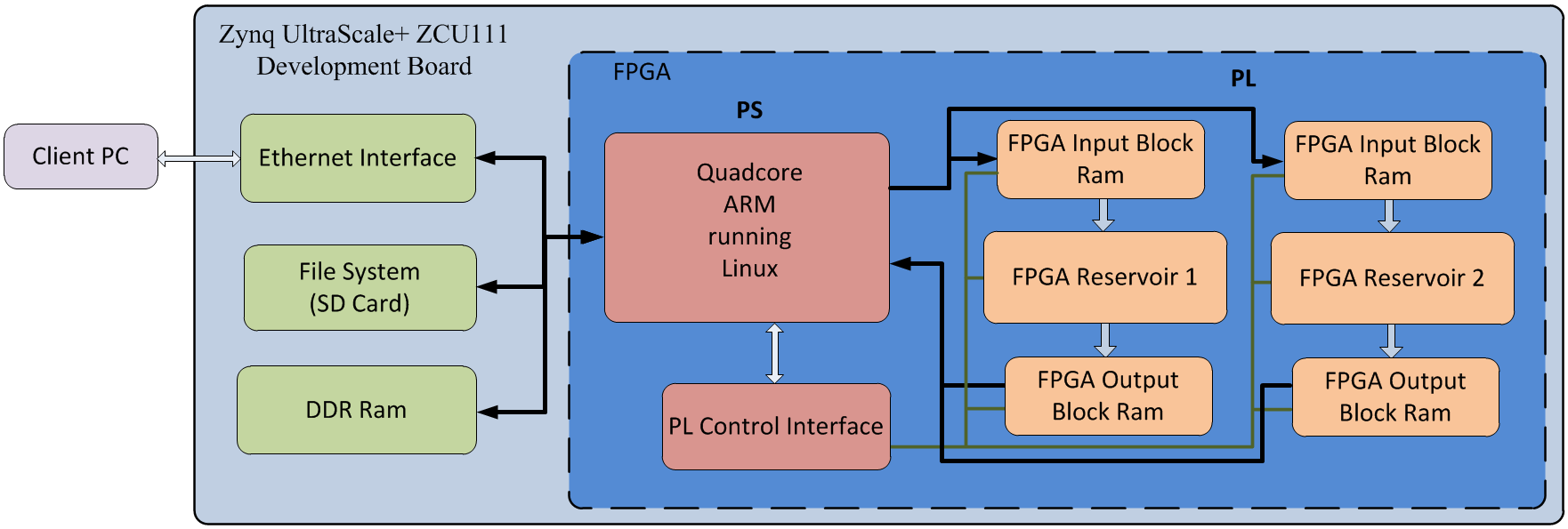}
  \captionof{figure}{FPGA dual loop block diagram }
  \label{fig:fig19}
\end{figure}

The Arm core performs the input mapping function on the sample data before it
 is broken into blocks and fed into the PL input block ram. The control interface
  is then used to configure the PL with the length of the burst length $M$
   and the number of sample sets in the input ram block. A state machine built into the PL controls processing the samples in the FPGA fabric. 
   The PS then reads vector $\ora{X_i}$ from the output FPGA block ram and performs the regression to calculate the total accuracy. See figure 20 for the full process flow.

   \begin{figure}[h!]
    \centering
    \includegraphics[width=14cm,height=7cm]{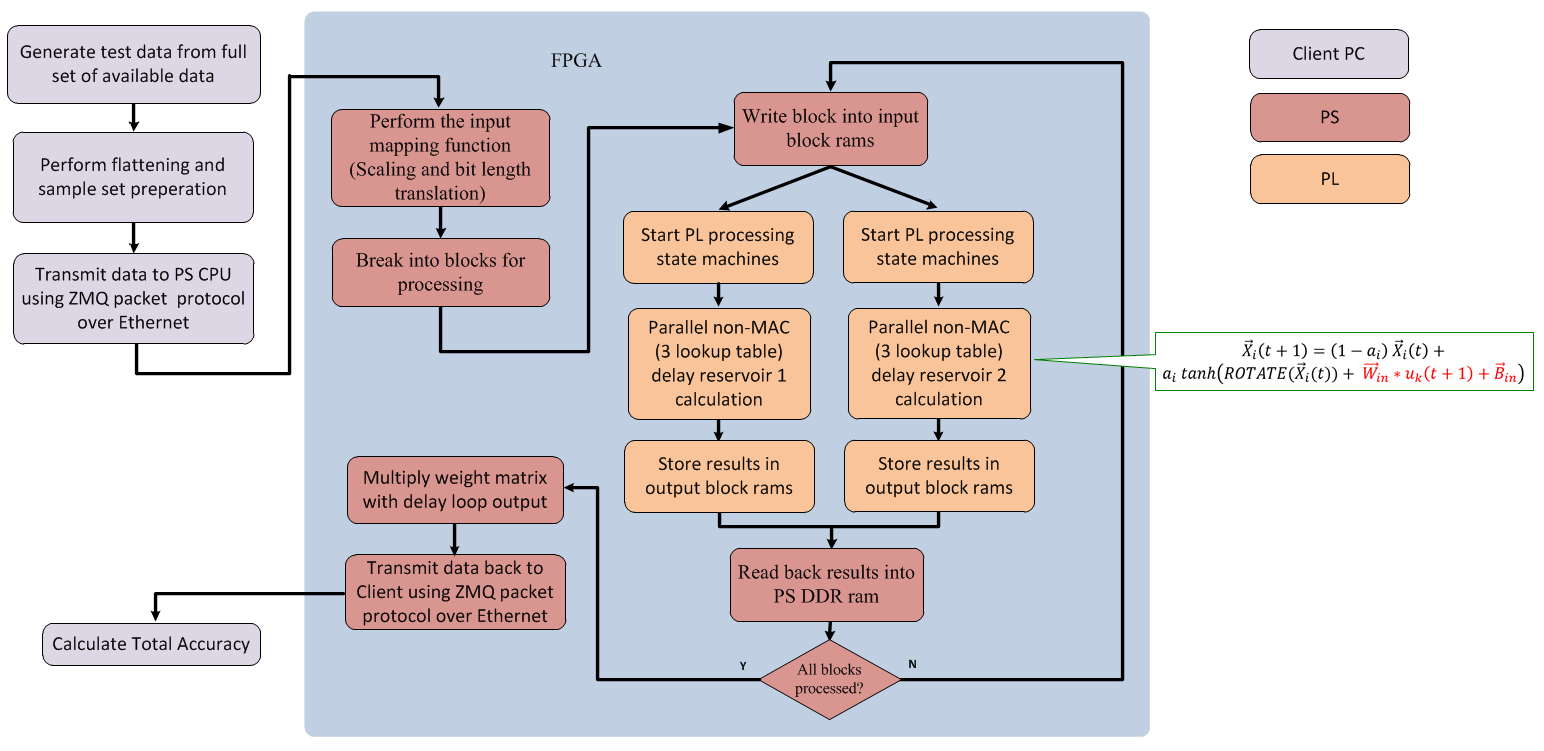}
    \captionof{figure}{FPGA process flow }
    \label{fig:fig20}
  \end{figure}

  The Zynq UltraScale+ XCZU28DR FPGA that we are using to develop the algorithms has a total of 1080 BRAM (Block Random Access Memory) resource blocks and 425,280 Look-Up Tables (LUT) resource blocks in its fabric. BRAMs are 36Kbit random access memory cells useful in building lookup tables and data buffers. They are one of four commonly identified components on an FPGA datasheet. The other three are Flip-Flops, Look-Up Tables (LUTs), and Digital Signal Processors (DSPs). In STARE, LUTs are used to implement the q-bit quantization for non-MAC computation.

  The current FPGA resource utilization for the STARE Single Delay-Loop Reservoir
   of dimensionality $N$ of 512-bit (including all support logic) is summarized in Table 1.

   \begin{figure}[h!]
    \centering
    \includegraphics[width=6cm,height=3cm]{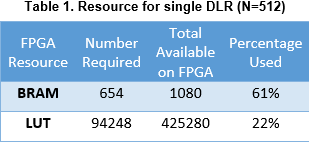}
    \label{fig:table1}
  \end{figure}

  The current FPGA resource utilization for the STARE Dual Delay-Loop Reservoir of dimensionality $N$ of 512-bit (including all support logic) is summarized in Table 2.

  \begin{figure}[h!]
    \centering
    \includegraphics[width=6cm,height=3cm]{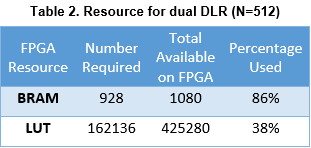}
    \label{fig:table2}
  \end{figure}

\subsection{ASIC}
This section investigates the future considerations required to transition from FPGA to an ASIC implementation and an estimate of the complexity.

\subsubsection{Types of ASIC Devices}

Implementation of the STARE architecture in microelectronics hardware can take into consideration multiple types of application specific integrated circuit (ASIC) devices. There are five classes of ASIC devices including application specific standard part (ASSP), programmable logic devices (PLD) and field programmable gate array (FPGA), structure ASIC, standard cell ASIC, and full custom ASIC. These classes of devices are listed in Table 3 in order of increasing cost, complexity, and performance. An ASSP is typically the lowest cost and risk, however this type of device is either general purpose or limited to type of operation, thus limiting overall performance. The full custom ASIC is the highest cost and risk, with the promise of highest performance with optimized power and area utilization.

  \begin{figure}[h!]
    \centering
    \includegraphics[width=16cm, height=8cm]{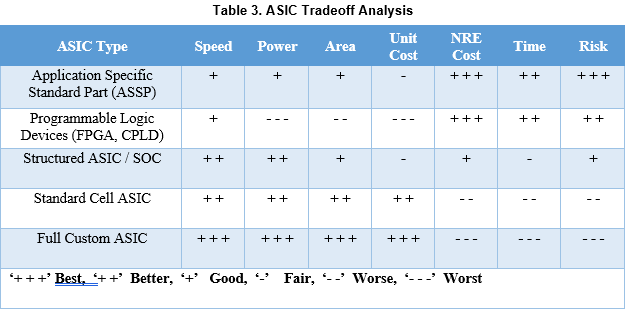} 
    \label{fig:table3}
  \end{figure}

  \subsubsection{STARE FPGA Utilization}

  The hardware implementation of the STARE reservoir loop is under development and the latest resource utilization estimates for the parallel implementation with a single reservoir loop of N=500 is provided in Table 4.

  \begin{figure}[h!]
    \centering
    \includegraphics[width=14cm, height=2.25cm]{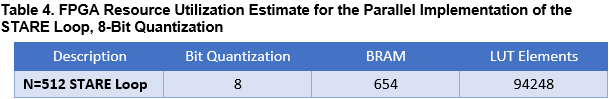} 
    \label{fig:table4}
  \end{figure}

  The resource utilization estimates in Table 5 have been updated for the Xilinx Zynq Ultrascale+ ZU28 and the Virtex Ultrascale+ VU19P.

  \begin{figure}[h!]
    \centering
    \includegraphics{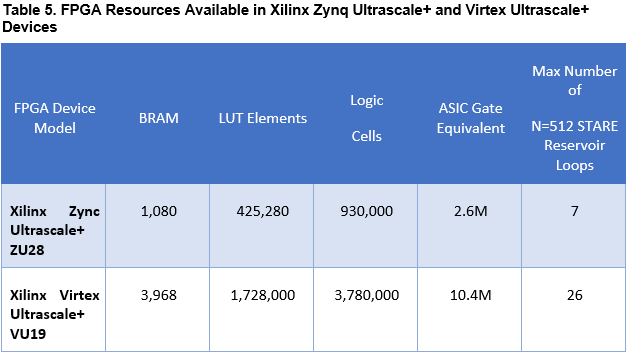}
    \label{fig:table5}
  \end{figure}

  \subsubsection{STARE ASIC Gate Count Estimation}

  The place and route of the STARE algorithm onto an FPGA does not fully utilize the logic and processing resources available in the FPGA. This creates difficulty extrapolating the resource estimates from the target FPGA to a standard cell ASIC implementation. The standard cell ASIC will have the flexibility advantage to place and route resources more efficiently. Guidelines for extrapolating the resource estimates include a conversion factor of one FPGA LUT to six ASIC gates. This conversion can range from three gates to eight gates and is based on the standard practice to define the measure of a single ASIC gate as a two input NAND logic operation, and the equivalent LUT resource utilization in an FPGA. The conversion for a 1024x8-bit BRAM is based on seven ASIC gates per memory bit. In order to convert the 8-bit fixed point DSP multiplier elements to ASIC gates, the synthesis constraints on the design tools were restricted to target FPGA LUT resources to avoid utilizing the FPGA DSP resources, thus providing an FPGA LUT and Register equivalent for the DSP multiplier, which is then converted to ASIC gates using the LUT 
  and Register conversion rates. The ASIC gate equivalent for the N=500 parallel STARE Loop and the N=20,000 are presented in Table 6.

  \begin{figure}[h!]
    \centering
    \includegraphics[width=12cm,height=4.6cm]{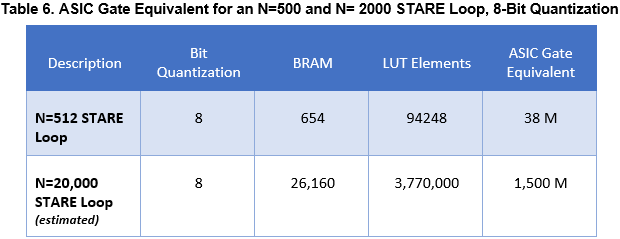} 
    \label{fig:table6}
  \end{figure}

  \subsubsection{STARE ASIC Architecture}

  In this section a proposed STARE ASIC architecture is presented for the development of an efficient structure for integrating a massively parallel implementation of the STARE algorithm. The basic building block is the STARE reservoir loop. The parallel hardware implementation of this loop is referred to as a STARE core and a grouping of 100 STARE cores as shown in Figure 21 is referred to as a Hyper Dimensional 100 (HD100) block. These blocks of cores will be tightly integrated to propagate input data samples, coefficient data values, intermediate data values, and output data values.

  \begin{figure}[h!]
    \centering
    \includegraphics{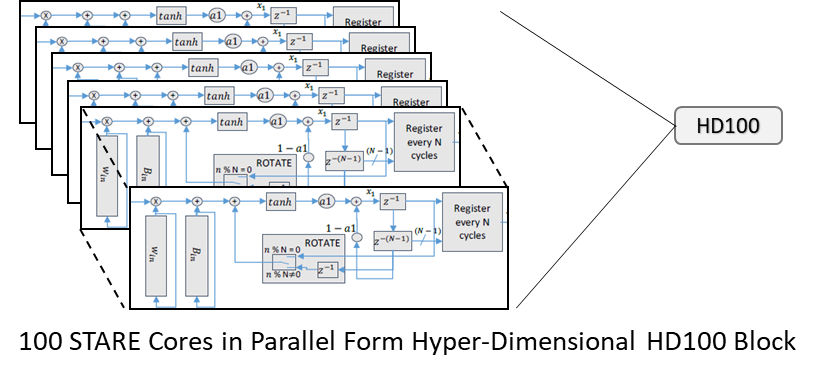} 
    \captionof{figure}{Hyper Dimensional 100 (HD100) Block of 100 STARE Cores }
    \label{fig:fig21}
  \end{figure}

  The next level of integration is a tiled structure of 20 HD100 blocks combined with a matrix multiplication unit, referred to as a Hyper Dimensional Reservoir Cluster (HDRC). High speed memory interfaces bring data samples, coefficients, and weights into the block, store intermediate data vectors, and store the output of the matrix multiplication unit. 
  
  The top level of a conceptual STARE ASIC architecture (see Figure 22) includes 20 HDRC blocks, a PCIe system interface, high bandwidth DRAM memory interfaces, and high-speed serial chip-to-chip interface implementing Interlaken or similar high bandwidth data interface between two STARE ASIC devices or between the STARE ASIC and a sensor chip such as a high definition, uncompressed video imager.

  \begin{figure}[h!]
    \centering
    \includegraphics{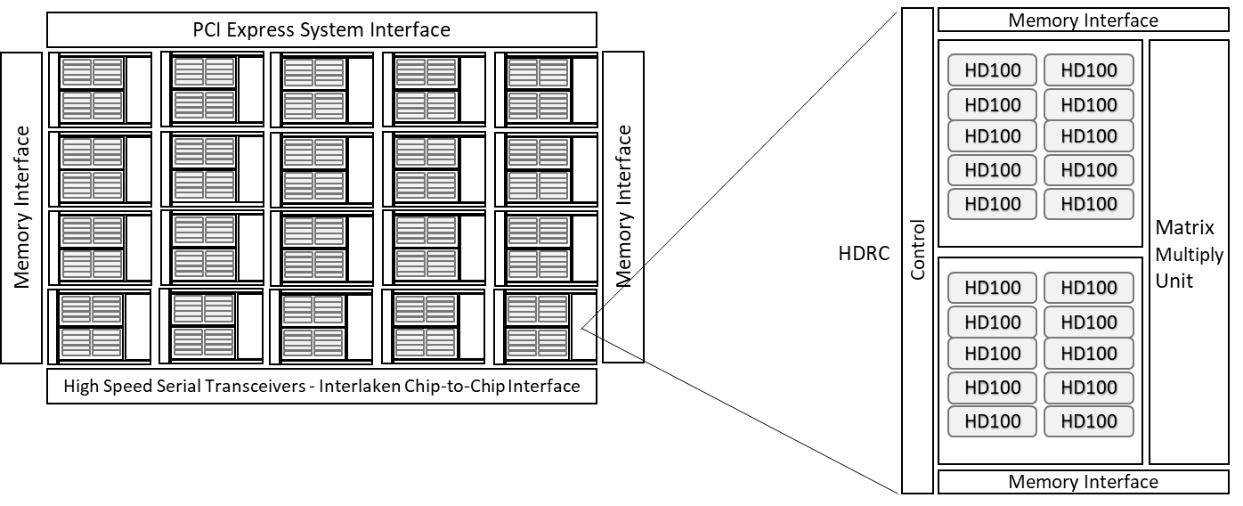} 
    \captionof{figure}{STARE ASIC Architecture with 20 Hyper Dimensional Reservoir Clusters }
    \label{fig:fig22}
  \end{figure}

  The STARE ASIC has 20 HDRCs, which contains a total of 20,000 STARE cores. With these cores operating at a nominal 1GHz clock speed the ASIC would perform 20 tera operations per second (an operation is the combined processing within one STARE core loop for this example). The estimated number of logic gates in the STARE ASIC is 61.9M gates.

  Using the Global Foundries 12nm fab process with a gate density of 8.5M gates/square mm, the STARE logic would have an estimated die size of 7.2 square mm (2.7 mm x 2.7mm). This estimate will increase with the addition of the matrix multiplication unit, system interface, memory interface, and serial interface. Relative to the NVIDIA, Intel, and Xilinx die sizes listed in Table 5, the 7.2 square mm STARE die area estimate is small and leaves significant room for the additional interfaces and circuitry and for the consideration of including an embedded processor, such as a multi-core ARM processor.

  \begin{figure}[h!]
    \centering
    \includegraphics[width=12cm,height=4cm]{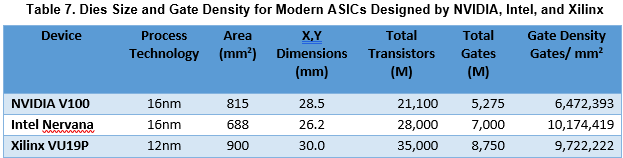} 
    \label{fig:table7}
  \end{figure}

  \section{Comparison of STARE and SoA Implementation}

  In this section, we compare STARE with state of the art (SOA) deep neural networks on the aforementioned applications. The following Tables summarize the comparison. Table 8 breaks down the comparison with respect to 3 different types of data: 1) Bird/drone classification, 2) Mackey Glass time series prediction, and 3) DeepSig modulation classification. Table 9 summarizes the comparison result for the latter 2 applications on the metrics including: accuracy, computation complexity, combined power and throughput and power, and training time on GPU. 
  
  From the comparison, STARE performance can be seen to be on par with ResNet for DeepSig data classification and outperforms LSTM performance for modulated Mackey Glass time series prediction. In both cases, STARE has significant advantage in training time and computation complexity.

  \begin{figure}[h!]
    \centering
    \includegraphics{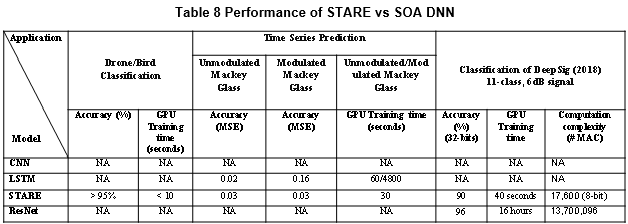}
    \label{fig:table8}
  \end{figure}

  \begin{figure}[h!]
    \centering
    \includegraphics[width=10cm,height=6cm]{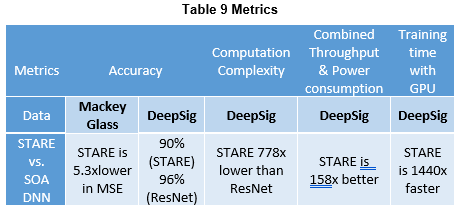}
    \label{fig:table9}
  \end{figure}

\section{Conclusion}

This paper describes a systematic approach towards building a new family of neural networks based on a delay-loop version of a reservoir neural network. The proposed Scaled-Time-Attention Robust Edge (STARE) network, architecture exploits high dimensionality to simplify training. It results in a shallow structure, is simpler to train, and is better suited for edge applications, including Internet of Things (IoT), due to its lower complexity and smaller footprint. STARE incorporates new AI concepts such as Attention and Context, and is best suited for temporal feature extraction and classification. We demonstrated that STARE is applicable to a variety of applications with improved performance and lower implementation complexity. In particular, we show a novel way of applying a dual-loop configuration to detection and identification of drone vs bird in a counter-UAS detection application by exploiting both spatial (video frame) and temporal (trajectory) information. We also demonstrated that the STARE performance approaches those of State-of-the-Art deep neural networks in classifying RF DeepSig data, and outperforms LSTM in a special case of Mackey Glass time series predictions. To illustrate feasibility of STARE for IoT applications, we implemented various STARE configurations in FPGA and demonstrated complete operation for counter UAS detection and identification. We also illustrated that the STARE approach is amenable to low-power, high-throughput implementation in an ASIC and provided gate count estimates for various configurations.

\section{Acknowledgement}

This paper is based upon work supported by the DARPA Artificial Intelligence Exploration program (PA-19-03-03) on HyDDENN. We thank Dr. Young-Kai Chen, Program Manager, for his vision on hyper-dimensional neural networks and his guidance of the program, Stefan Westberg (AFRL) for his suggestions on visual processing related to counter UAS detection, and Greg Jones for many technical discussions. The views, opinions and/or findings expressed are those of the authors and should not be interpreted as representing the official views or policies of DARPA or the U.S. Government.



\bibliographystyle{unsrt}  
\bibliography{references}






\end{document}